# TopRank+: A Refinement of TopRank Algorithm

Dr. Victor H. de la Peña[1]     Haolin Zou[2]

## Abstract

Online learning to rank is a core problem in machine learning. In Lattimore et al. (2018), a novel online learning algorithm was proposed based on topological sorting. In the paper they provided a set of self-normalized inequalities (a) in the algorithm as a criterion in iterations and (b) to provide an upper bound for cumulative regret, which is a measure of algorithm performance. In this work, we utilized method of mixtures and asymptotic expansions of certain implicit function to provide a tighter, iterated-log-like boundary for the inequalities, and as a consequence improve both the algorithm itself as well as its performance estimation.

## 1. Introduction

Online learning to rank is a core problem in recommender systems and web search engines. The goal is to learn an ordered subset of total items to maximize user satisfaction, based on sequential feedback, usually in the form of user clicks. A click model is a stochastic model on user behavior of clicking [1]. Cascade model

[1] Columbia University, New York, NY

[2] Free Researcher, New York, NY

(CM) [2] and position-based model (PBM) [3] are two popular click models. CM assumes that a user would examine items in given order and click the first item they find attractive, while in PBM the probability that user would click on one item depends on its attractiveness as well as its position in the recommendation.

Zoghi et al. [8] proposed a new online ranking algorithm based on a generalized click model nesting both CM and PBM. Lattimore et al. [4] moved further by generalizing the click model of Zoghi et al. [8] and proposed a new online ranking algorithm called TopRank, which is based on topological sorting. The general idea of the algorithm is to maintain a directed acyclic graph (DAG) of item pairwise attractiveness, and add edges sequentially to the graph if certain confidence criterion is met. The criterion can be expressed by a boundary-crossing inequality, which is also a key to estimate the upper bound of overall performance, measured by cumulative regret.

We refined the above inequalities using method of mixtures proposed by de la Peña et. al. [6][7]. The refinement results in an iterated-log inequality in both algorithm criterion and regret boundary, improving both the algorithm itself as well as estimation of its performance.

## 2. TopRank

### 2.1. Algorithm

TopRank is an online learning ranking algorithm proposed by Lattimore et al. [4]. Based on a group of general assumptions on user behavior (called click model), where the problem of finding an optimal list ordered by attractiveness can be posed as a sorting problem with noisy feedback. TopRank algorithm works as follows:

Assume there are L items to be ranked, indexed by i=1,2, …, L. The goal is to rank these L items by attractiveness to users. Furthermore, only the first K most attractive items are of interest. Information is provided sequentially in the form of user clicks. To be specific, define $C_{ti}=1$ if user clicks on item i in round t=1, 2, …, n. For each round, the algorithm provides a permutation $A_t$ of all items based on previous clicks, display the first K items to the user, and collect new clicks. Ideally the permutation becomes closer to the 'optimum' ranking after each iteration.

Assumptions on the probability $\Pr(C_{ti} = 1)$ is called click model. To be specific, let $A_t(k)$ be the kth item in permutation $A_t$:

$$v(A_t, k) := \Pr\left(C_{tA_{t(k)}} = 1\right) \tag{1}$$

Let $v(a, k) = \alpha(a(k)) * \chi(a, k)$, where $\alpha$ is attractiveness function of items, $\chi$ is the probability that user checks kth item in permutation a. An optimal permutation $a^*$ satisfies:

$$\alpha(a^*(1)) \geq \alpha(a^*(2)) \geq \cdots \geq \alpha(a^*(K)) \tag{2}$$

For simplicity and without loss of generality, it can be assumed that $a^* = [1, 2, \ldots, L]$

Function $v$ follows several assumptions:

**Assumption 1.** $v(a, k) = 0 \; for \; all \; k > K$

**Assumption 2.** *Let $a^*$ be an optimal permutation. Then* $\max_a \sum_{k=1}^{K} v(a, k) = \sum_{k=1}^{K} v(a^*, k)$

**Assumption 3:** *Let i and j be items with $\alpha(i) \geq \alpha(j)$. For any permutation a, let a' be the permutation that exchanges only item i and j in a. Then we have*:

$$v(a, a^{-1}(i)) \geq \frac{\alpha(i)}{\alpha(j)} v(a', a^{-1}(i))$$

**Assumption 4:** *For any permutation a and optimal $a^*$ with $\alpha(a(k)) = \alpha(a^*(k))$, we have*

$$v(a, k) \geq v(a^*, k)$$

It can be proven that such set of assumptions nests most commonly used click models, including cascade model (CM), position-based model (PBM), and even a more general model proposed by Zoghi et al. (2007) [8]

Define cumulative regret as the cumulative difference between expected number of clicks on $A_t$: and global 'optimal' ranking $a^*$:

$$R_n = n * \sum_{k=1}^{K} v(a^*, k) - E\left(\sum_{t=1}^{n}\sum_{i=1}^{L} C_{ti}\right) = \max_{a} E\left(\sum_{t=1}^{n}\sum_{i=1}^{L} v(a,k) - v(A_t, k)\right)$$

Based on Kahn's topological sorting, the TopRank algorithm sequentially adds new edges to a Directed Acyclic Graph (DAG) $G$.

---

**TopRank Algorithm**

Set $G_1 = \emptyset$ and $c = 4\sqrt{\frac{2}{\pi}} / \text{erf}(\sqrt{2})$

For t = 1, 2, …, n:

    Do Kahn's topological sorting based on graph $G_t$ and return permutation $A_t$

    Collect user clicks $C_{ti}$ for $i \in [L]$

    For i, j $\in [L]^2$:

$$U_{tij} = \begin{cases} C_{ti} - C_{tj}, & \text{if } i,j \in P_{td} \text{ for some } d \\ 0, & \text{else} \end{cases}$$

$$S_{tij} = \sum_{s=1}^{t} U_{sij}$$

$$N_{tij} = \sum_{s=1}^{t} |U_{sij}|$$

$$G_{t+1} = G_t \cup \left\{(j,i) : S_{tij} \geq \sqrt{2N_{tij} \log(\frac{c}{\delta}\sqrt{N_{tij}})} \text{ and } N_{tij} > 0\right\}$$

The inequality in the last line is of our concern here: it controls when we have enough 'confidence' that item i is more attractive than item j. Let's restate it here:

$$S_{tij} \geq \sqrt{2N_{tij} \log\left(\frac{c}{\delta}\sqrt{N_{tij}}\right)} \text{ and } N_{tij} > 0 \tag{3}$$

The refinement is mainly made on this inequality.

## 2.2. Regret Analysis

**Theorem 1.** *Let function v satisfy assumptions 1 - 4 and* $\alpha(1) > \alpha(2) > \cdots > \alpha(L)$. *Let* $\Delta_{ij} = \alpha(i) - \alpha(j)$ *and* $\delta \in (0,1)$. *The n-step regret is bounded from above as*

$$R_n \leq \delta n K L^2 + \sum_{j=1}^{L} \sum_{i=1}^{\min\{K,j-1\}} \left(1 + \frac{6(\alpha(i) + \alpha(j)) \log\left(\frac{c\sqrt{n}}{\delta}\right)}{\Delta_{ij}}\right) \tag{4}$$

*Furthermore,*

$$R_n \leq \delta n K L^2 + KL + \sqrt{4K^3 Ln * \log\left(\frac{c\sqrt{n}}{\delta}\right)} \tag{5}$$

.

By choosing $\delta = \frac{1}{n}$, we have $R_n = O(\sum_{j=1}^{L} \sum_{i=1}^{\min\{K,j-1\}} (\frac{\alpha(i)\log(n)}{\Delta_{ij}}))$ and $R_n = O(\sqrt{K^3 Ln * \log(n)})$

To prove Theorem 1, six lemmas were proposed as below: (without proof here)

**Lemma 1.** *Let i and j satisfy* $\alpha(i) \geq \alpha(j)$ *and* $d \geq 1$. *Let* $I_{td} = [\sum_{c=1}^{d-1}|P_{tc}| + 1, \sum_{c=1}^{d}|P_{tc}|]$ *be the slots of items in* $P_{td}$. *Let* $M_t$ *be the number of blocks in round t. On the event that* $i, j \in P_{sd}$ *and* $d \in [M_s]$ *and* $U_{sij} \neq 0$, *the following hold almost surely:*

(a) $E_{s-1}[U_{sij}|U_{sij} \neq 0] \geq \frac{\Delta_{ij}}{\alpha(i)+\alpha(j)}$ 　　　　(b) $E_{s-1}[U_{sji}|U_{sji} \neq 0] \leq 0$

**Lemma 2.** Let $F_t$ be a failure event that $\exists\, i \neq j$ and $s < t$ s.t. $N_{sij} > 0$ and

$$\left| S_{sij} - \sum_{u=1}^{s} E_{u-1}[U_{sij}|U_{sij} \neq 0]\,|U_{uij}| \right| \geq \sqrt{2 N_{sij} \log\left(\frac{c\sqrt{N_{sij}}}{\delta}\right)} \qquad (6)$$

Then it holds that $Pr(F_n) \leq \delta L^2$

**Lemma 3.** On the event $F_t^c$ it holds that $\forall\, i < j, (i,j) \notin G_t$. Equivalently speaking, when $F_t$ doesn't happen, the algorithm never makes a wrong decision in comparing item i and j.

**Lemma 4.** Let $V_{td}$ be the most attractive item in $P_{td}$. Then on event $F_t^c$ it holds that $V_{td} \leq 1 + \sum_{c<d} |P_{td}|, \forall d \in [1, M_t]$.

**Lemma 5.** On the event $F_n^c$ and $\forall\, i < j$ it holds that

$$S_{nij} \leq 1 + \frac{6(\alpha(i)+\alpha(j))}{\Delta_{ij}} \log\left(\frac{c\sqrt{n}}{\delta}\right) \qquad (7)$$

**Lemma 6.** Let $\{X_t\}$ be an adapted process with $X_t \in \{-1,0,1\}$ and $\mu_t = E_{t-1}[X_t|X_t \neq 0]$. Then with $S_t = \sum_{s=1}^{t}(X_s - \mu_s|X_s|)$ and $N_t = \sum_{s=1}^{t} |X_s|$, we have:

$$Pr\left(\exists\, t > 0, |S_t| \geq \sqrt{2 N_t \log\left(\frac{c\sqrt{N_t}}{\delta}\right)} \text{ and } N_t > 0\right) \leq \delta, \text{ where } c = \frac{4\sqrt{\frac{2}{\pi}}}{erf(\sqrt{2})} \text{ is constant.} \qquad (8)$$

Note that Lemma 6 is independent from the algorithm and has a close relationship with boundary crossing probabilities (H. Robbins and D. Siegmund, 1970 [7]). In the following chapters, we use method of mixtures proposed by de la Pena et al. (2004) [6], it is possible to tighten the boundary of Lemma 6 and refine the algorithm. The refined inequalities have an interesting relationship with Law of Iterated Logarithms (LIL).

# 3. Method of Mixtures

Assume $\left\{e^{\lambda A_t - \frac{\lambda^2}{2}B_t^2}\right\}_{t\geq 0}$ is a supermartingale with mean $\leq 1$ for $0 \leq \lambda \leq \lambda_0$ and $A_0 = 0$.

Let F be a finite and nontrivial positive measure on $(0, \lambda_0)$. Let

$$\Psi(u, v) = \int_0^{\lambda_0} \exp\left(\lambda u - \frac{\lambda^2 v}{2}\right) dF(\lambda) \tag{9}$$

Since $\Psi$ is continuous and monotonously increasing w.r.t u, given $c > 0$ and $v > 0$, the equation

$$\Psi(u, v) = c \tag{10}$$

has a unique solution

$$u = \beta_F(v, c)$$

It is easy to show that

$$\Pr(\exists t \text{ s.t. } A_t \geq \beta_F(B_t^2, c)) = \Pr(\exists t \text{ s.t. } \Psi(A_t, B_t^2) \geq c) \leq \frac{F(0, \lambda_0)}{c} \tag{11}$$

using the fact that $\beta_F$ is monotone to v and applying Doob's inequality (see also Freedman (1975) [9]) to the

$\Psi(A_t, B_t^2)$.

**Lemma (Robbins and Siegmund) [7]:**

For $\lambda \in (0, e^{-e})$, let

$$dF(\lambda) = \frac{1}{\lambda \log\frac{1}{\lambda}\left(loglog\frac{1}{\lambda}\right)^2} d\lambda \tag{12}$$

Then $F(0, e^{-e}) = 1$ and for fixed $c > 0$,

$$\beta_F(v, c) = \sqrt{2v\left[loglog(v) + \frac{5}{2}logloglog(v) + \log\left(\frac{c}{2\sqrt{\pi}}\right) + o(1)\right]} \tag{13}$$

as $v \to \infty$. Furthermore,

$$\overline{\lim} \frac{A_t}{\sqrt{2B_t^2 \log\log B_t^2}} \leq 1 \tag{14}$$

Using (11) and (13), we can modify (18) in lemma 6:

$$\Pr\left(\exists t > 0 \ s.t. \ |S_t| \geq \beta_F\left(N_t, \frac{1}{2\delta}\right) \text{ and } N_t > 0\right) \leq \delta \tag{15}$$

To get a closed form inequality, we need a closed form boundary for $\beta_F\left(N_t, \frac{1}{2\delta}\right)$.

By (13),

$$\beta_F\left(N_t, \frac{1}{2\delta}\right) = \sqrt{2N_t\left[\log\log(N_t) + \frac{5}{2}\log\log\log(N_t) + \log\left(\frac{1}{4\delta\sqrt{\pi}}\right) + o(1)\right]} \tag{16}$$

Note that $\lim_{t \to \infty} N_t \to \infty, a.s.$ can be guaranteed by the design of algorithm.

One attempt is to bound the $o(1)$ term in (16) from above. Suppose $C_0$ is a constant upper bound for the $o(1)$ term.

$$\beta_F\left(N_t, \frac{1}{2\delta}\right) \leq \sqrt{2N_t\left[\log\log(N_t) + \frac{5}{2}\log\log\log(N_t) + C_1(\delta)\right]} \tag{17}$$

Where $C_1(\delta) = \log\left(\frac{1}{4\delta\sqrt{\pi}}\right) + C_0$. Such $C_1$ should exist but it is difficult to write a formula according to Robbins and Siegmund (1970) [7].

Another attempt is to control the constants and triple logarithm term in (16) by a double log:

$$\beta_F\left(N_t, \frac{1}{2\delta}\right) \leq \sqrt{(2 + C_2(\delta))N_t \log\log(N_t)} \tag{18}$$

(17) and (18) lead to two versions of modifications on (8) in Lemma 6, which results in further refinements of (3) in algorithm; (6) in Lemma 2; (7) in lemma 5; and (4), (5) in Theorem 1. The details are in the next chapter.

# 4. Refinements

## 4.1. Tighter but complex version

First let's restate (17):

$$\beta_F\left(N_t, \frac{1}{2\delta}\right) \leq \sqrt{2N_t\left[loglog(N_t) + \frac{5}{2}logloglog(N_t) + C_1(\delta)\right]} \qquad (17)$$

Insert into (11) we have:

**Lemma 6'.** (assumptions are similar to Lemma 6)

$$\Pr\left(\exists t > 0, |S_t| \geq \sqrt{2N_t\left[loglog(N_t) + \frac{5}{2}logloglog(N_t) + C_1(\delta)\right]} \text{ and } N_t > 0\right) \leq \delta.$$

Similarly, Lemma 2 still holds if we change the definition of $F_t$:

**Lemma 2'.** Let $F_t$ be a failure event that $\exists i \neq j$ and $s < t$ s.t. $N_{sij} > 0$ and

$$\left|S_{sij} - \sum_{u=1}^{s} E_{u-1}\left[U_{sij}|U_{sij} \neq 0\right]|U_{uij}|\right| \geq \sqrt{2N_t\left[loglog(N_t) + \frac{5}{2}logloglog(N_t) + C_1(\delta)\right]} \qquad (19)$$

Then it holds that $\Pr(F_n) \leq \delta L^2$

Lemma 3 still holds if we change criteria (3) in the algorithm into:

$$S_{tij} \geq \sqrt{2N_t\left[loglog(N_t) + \frac{5}{2}logloglog(N_t) + C_1(\delta)\right]} \text{ and } N_{tij} > 0 \qquad (20)$$

Similarly Lemma 5 can be modified:

**Lamma 5'.**

$$S_{nij} \leq 1 + \frac{6(\alpha(i)+\alpha(j))}{\Delta_{ij}} (loglog(n) + \frac{5}{2} logloglog(n) + C_1(\delta)) \tag{21}$$

Lemma 1 and Lemma 4 are invariant to these inequalities thus still hold.

Finally (4) and (5) in Theorem 1 can be refined as:

**Theorem 1'.** (assumptions similar as before)

$$R_n \leq \delta nKL^2 + \sum_{j=1}^{L} \sum_{i=1}^{\min\{K,j-1\}} \left(1 + \frac{6(\alpha(i)+\alpha(j))\left(loglog(n)+\frac{5}{2}logloglog(n)+C_1(\delta)\right)}{\Delta_{ij}}\right) \tag{22}$$

$$R_n \leq \delta nKL^2 + KL + \sqrt{4K^3Ln * \left(loglog(n) + \frac{5}{2} logloglog(n) + C_1(\delta)\right)} \tag{23}$$

## 4.2. Simpler version

Similar as in 3.1 but use (18) instead of (17) results in another version of modifications:

**Lemma 6".** (assumptions are similar to Lemma 6)

$$\Pr\left(\exists t > 0, |S_t| \geq \sqrt{(2+C_2(\delta))N_t loglog(N_t)} \text{ and } N_t > 0\right) \leq \delta.$$

Similarly, Lemma 2 still holds if we change the definition of $F_t$:

**Lemma 2".** Let $F_t$ be a failure event that $\exists\, i \neq j$ and $s < t$ s.t. $N_{sij} > 0$ and

$$\left| S_{sij} - \sum_{u=1}^{s} E_{u-1}\left[U_{sij}|U_{sij} \neq 0\right] |U_{uij}| \right| \geq \sqrt{(2+C_2(\delta))N_t loglog(N_t)} \tag{19}$$

Then it holds that $\Pr(F_n) \leq \delta L^2$

Lemma 3 still holds if we change criteria (3) in the algorithm into:

$$S_{tij} \geq \sqrt{(2+C_2(\delta))N_t loglog(N_t)} \text{ and } N_{tij} > 0 \tag{20}$$

Similarly Lemma 5 can be modified:

**Lamma 5".**

$$S_{nij} \leq 1 + \frac{\left(1+2\sqrt{2+C_2(\delta)}\right)(\alpha(i)+\alpha(j))}{\Delta_{ij}} loglog(n) \qquad (21)$$

Lemma 1 and Lemma 4 are invariant to these inequalities thus still hold.

Finally (4) and (5) in Theorem 1 can be refined as:

**Theorem 1".** (assumptions similar as before)

$$R_n \leq \delta nKL^2 + \sum_{j=1}^{L} \sum_{i=1}^{\min\{K,j-1\}} \left(1 + \frac{\left(1+2\sqrt{2+C_2(\delta)}\right)(\alpha(i)+\alpha(j))loglog(n)}{\Delta_{ij}}\right) \qquad (22)$$

$$R_n \leq \delta nKL^2 + KL + \sqrt{2(2+C_2(\delta))K^3 Ln * loglog(n)} \qquad (23)$$

# 5. Conclusion

In this work we proposed a tighter boundary for cumulative regret of TopRank algorithm, and refined the key criterion accordingly. The improvement is asymptotic, namely when t and $N_t \to \infty$, the complexity is reduced to $\sqrt{n * loglog(n)}$. A next work can focus on finding a closed form of $C_1(\delta)$ and $C_2(\delta)$ to finish the discussion.